# Introduce the Result Into Self-Attention

Chengcheng Ye
Jilin University
netyuki@163.com

**Abstract**——Traditional self-attention mechanisms in convolutional networks tend to use only the output of the previous layer as input to the attention network, such as SENet, CBAM, etc. In this paper, we propose a new attention modification method that tries to get the output of the classification network in advance and use it as a part of the input of the attention network. We used the auxiliary classifier proposed in GoogLeNet to obtain the results in advance and pass them into attention networks. we added this mechanism to SE-ResNet for our experiments and achieved a classification accuracy improvement of at most 1.94% on cifar100.

## 1 Introduction

Various attention mechanisms have been widely used in convolutional networks with remarkable results. For example, Squeeze-and-Excitation attention networks [1] implement channel attention with a relatively small computational overhead. The later CBAM [2], coord attention [3], etc. implemented channel attention and spatial attention. A common feature of these attention mechanisms is that they only use the output of the previous layer as the input to the attention network. Theoretically, the more useful information passed into the attention network, the better it is for the attention network to give a reasonable weight distribution, and for the classification network, there is no more useful information than the prediction results given by the model. Accordingly, this paper proposes a new way to improve the attention module by combining the prediction results with the original attention network input to form a new attention module input.

Obviously, the improvement can simply be added to all current attention network methods that use the upper layer as input, and we first experimented on the widely used Squeeze-and-Excitation attention. The experimental network is SE-ResNet, and the experimental dataset is cifar100. The comparison of the experimental results before and after the network modification using auxiliary classifier is shown in Table 3.

Using the auxiliary classifier approach, the loss weights of the auxiliary classifier can have a significant impact on the final results, see Section 5.4 for a discussion of this. Also, the auxiliary classifier itself has a certain boost to the result, See Section 5.1 for a comparison test used to explore how much of auxiliary classifiers contribute to the total accuracy improvement.

## 2 Introduce The Result Into The Attention

Let's start by reviewing the Squeeze-and-Excitation operation of SENet. First denote the output result after a certain layer of convolution by $U=[u_1, u_2, \cdots, u_C]$, $U \in R^{H \times W \times C}$, Let one of the channels be $u_c$, $u_c \in R^{H \times W}$. Then the result after the squeeze operation on U is $z \in R^C$. Let one of the element be $z_c$, which is calculated as follows:

$$z_c = \frac{1}{H \times W} \sum_{i=1}^{H} \sum_{j=0}^{W} u_c(i,j) \tag{1}$$

That is, the global average pooling is performed for each channel. After obtaining z we use z as input and perform the excitation operation. The formula is as follows.

$$s = \sigma(W_2 \delta(W_1 z)) \tag{2}$$

where $\delta$ denotes the ReLU function, $W_1 \in R^{\frac{C}{r} \times C}$, $W_2 \in R^{C \times \frac{C}{r}}$, r is the reduction ratio. $\sigma$ is the sigmoid function. In brief, the excitation operation here is to input z into a fully connected layer $W_1$ and then pass through ReLU, then pass through a fully connected layer $W_2$, and finally pass through a sigmoid function to obtain $s \in R^C$. After obtaining s, U is rescaled with s to obtain the result after applying attention. The formula is as follows.

$$X' = F_{scale}(u_c, s_c) = s_c u_c \tag{3}$$

X' is the output after a Squeeze-and-Excitation.

Next, we describe how to introduce the results of auxiliary classifier in the Squeeze-and-Excitation operation. First we denote the result obtained from the auxiliary classifier as T. Let the number of categories be n, then $T \in R^n$. After obtaining z from Eq. (1), we concatenate z with T to obtain a new attention input z':

$$z' = F_{cat}(T, z) \tag{4}$$

The rest of the operation remains the same as before:

$$s = \sigma(W_2 \delta(W_1 z')) \tag{5}$$

$$X' = F_{scale}(u_c, s_c) = s_c u_c \tag{6}$$

See Figure 1 for a comparison of the structure before and after modification.

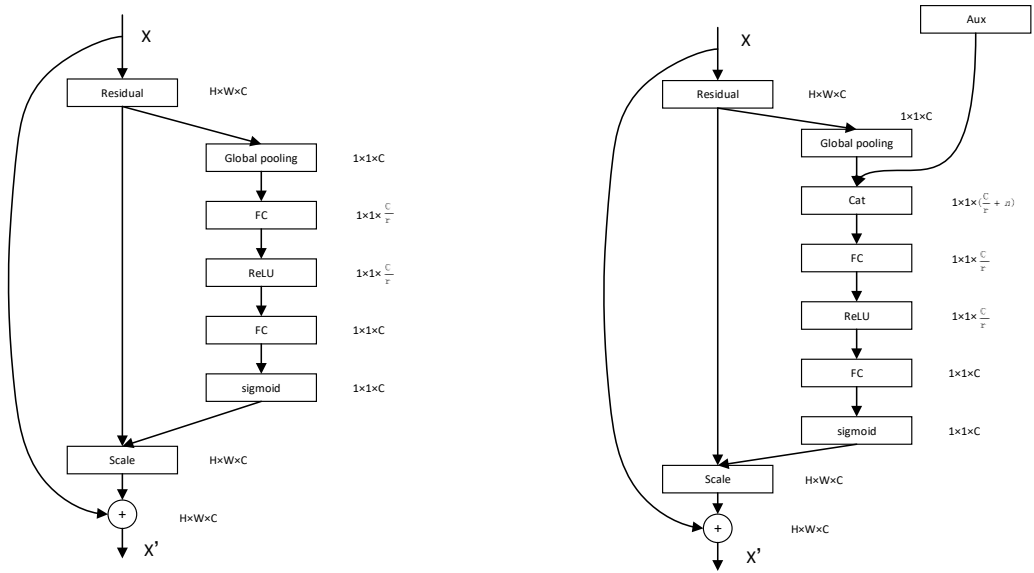

Fig.1. The left is the original SE-ResNet module, and the right is the module after adding auxiliary classifier, where n indicates the dimension of aux output

# 3 Model Structure, Number Of Parameters And Computational Complexity

## 3.1 Model Structure

We first did experiments on SE-ResNet34, SE-ResNet50, and SE-ResNet101, Let's name the added networks as SE-R-ResNet34, SE-R-ResNet50, and SE-R-ResNet101.. The dataset is cifar100. In order to make SE-ResNet better adapted to small data sets, we have made a small modification to it: Change the size of the first convolution kernel from 7*7 to 3*3, set stride=1 and padding=1. Remove one of the Max pooling operations behind this layer. As we know, the middle of SE-ResNet can be divided into four stages, each of which consists of repeating Blocks. We add auxiliary classifiers at the end of the second and third stages, and introduce the respective output results into each block of the subsequent stages as part of the attention input. The structure can be seen in Figure 2.

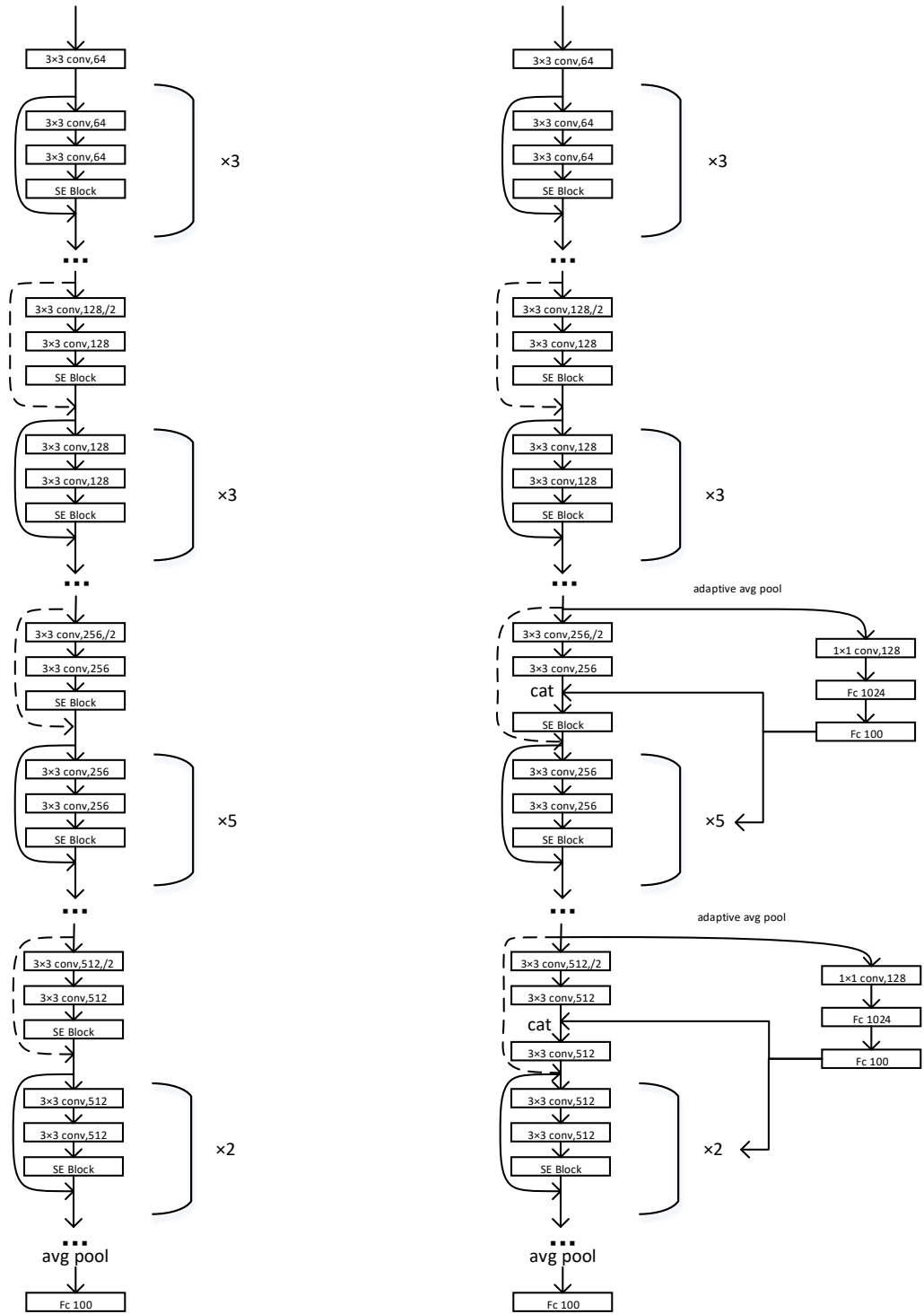

Fig. 2. SE-ResNet34 on the left (the first layer is modified and the subsequent maximum pooling is removed) and SE-R-ResNet34 on the right. (relu and batchnorm are omitted in the figure)

The structure of the Auxiliary classifier is consistent with that proposed in GoogLeNet [6], specifically, it consists of an adaptive average pooling operation, a convolutional layer, and two fully connected layers with a dropout operation between the two fully connected layers. The losses of each of the two auxiliary classifiers are multiplied by a weight added to the loss of the main network; here, we denote the auxiliary classifier near the input layer as aux1 and the other one as aux2, and the corresponding weights multiplied with it are denoted as lossW1 and lossW2.

## 3.2 Number Of Parameters

The increase in the number of parameters in the modified model is almost fixed compared with the original SE-ResNet, because the main increase in parameters comes from the fully connected layer of the two auxiliary classifiers, and the increase in parameters in the attention part is almost negligible. Therefore the percentage increase in the number of parameters is smaller when the number of parameters is larger for the model,. As shown in Table 1, SE-ResNet101, for example, increases the number of parameters by approximately 9.6%. More detailed results are shown in Table 1. all results in Table 1 were obtained at r=8.

| Net | Number of Parameters | Net | Number of Parameters |
|---|---|---|---|
| SE-ResNet34 | 21.655M | SE-R-ResNet34 | 26.145M |
| SE-ResNet50 | 28.779M | SE-R-ResNet50 | 33.531M |
| SE-ResNet101 | 52.273M | SE-R-ResNet101 | 57.243M |

Table1

## 3.3 Computational Complexity

The MACs of the three SE-ResNet modifications changed very little from the previous ones, increasing by 0.52%, 0.61%, and 0.32%, respectively. Because of the cifar100 dataset used for the experiments, the tensor dimensions of the inputs used to calculate the MACs are [1,3,32,32], as shown in Table 2. in addition, the tables are all obtained by the thop tool under the setting of r=8.

| Net | MACs | Net | MACs |
|---|---|---|---|
| SE-ResNet34 | 1.163G | SE-R-ResNet34 | 1.169G |
| SE-ResNet50 | 1.315G | SE-R-ResNet50 | 1.323G |
| SE-ResNet101 | 2.538G | SE-R-ResNet101 | 2.546G |

Table2

# 4 Experiments

SE-R-ResNet has a more significant improvement in classification accuracy compared to the original SE-ResNet, and we will show this result through experiments in this section.

The experimental network here adopts a structure similar to that described in fig2, i.e., the two auxiliary classifiers are added after the second stage and the third stage, and the dropout rate of the auxiliary classifier is 0.7. Unless otherwise specified, all networks were run with r = 8 and all used the same training and data augmentation strategies. The training strategies we used were: init lr = 0.1 divide by 5 at 60th, 120th, 160th epochs, train for 200 epochs with batchsize 128 and weight decay 5e-4, Nesterov momentum of 0.9 [4]. A warm up strategy was adopted in the first epoch [5], linearly increasing the learning rate gradually to 0.1. The data augmentation strategies are: random cropping with padding 4, and perform random horizontal flipping, random rotation (-15°,15°). Each input image is normalised through mean RGB-channel subtraction.

The experimental results are shown in Table 3. All three networks are improved compared with the original, among which, SE-R-ResNet101 achieves an accuracy improvement of 1.94, which is obvious.

The most appropriate lossW of SE-R-ResNet differs for different depths, for example, SE-R-ResNet34 is the loss of both aux multiplied by 0.4 added to the loss of the main network, while the loss of aux1 of SE-R-ResNet101 is multiplied by 0.3 and the loss of aux2 multiplied by 0.2 added to the loss of the main network.

| Net | Accuracy | Net | Accuracy | lossW1 | lossW2 |
|---|---|---|---|---|---|
| SE-ResNet34 | 78.12 | SE-R-ResNet34 | 78.98 | 0.4 | 0.4 |
| SE-ResNet50 | 78.22 | SE-R-ResNet50 | 79.96 | 0.3 | 0.3 |
| SE-ResNet101 | 78.54 | SE-R-ResNet101 | 80.48 | 0.3 | 0.2 |

Table3

Figure 3, 4, 5 shows in detail the error rate change curves of the network during training before and after the modification, and it can be observed that the improvement has produced effects throughout the training process.

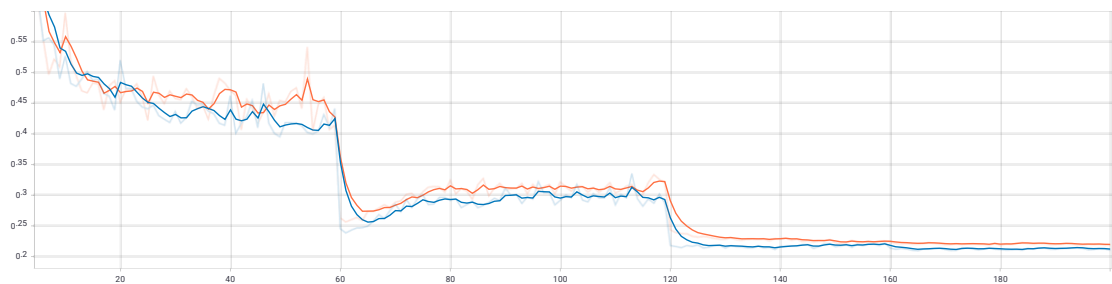

Fig. 3.  SE-ResNet34(orange)  SE-R-ResNet34(blue)

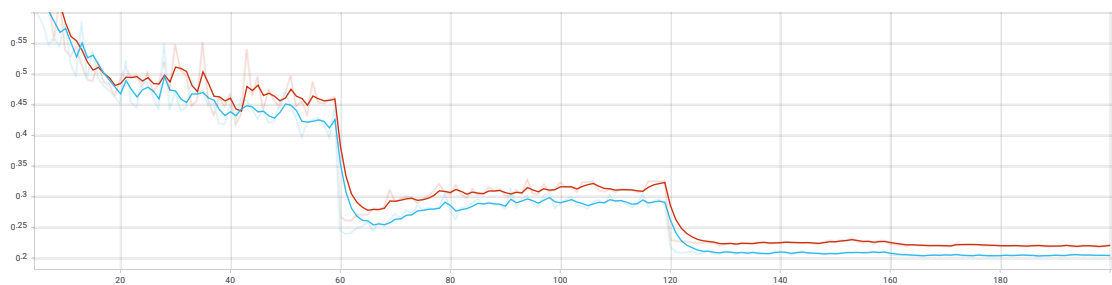

Fig. 4.  SE-ResNet50(orange)  SE-R-ResNet50(blue)

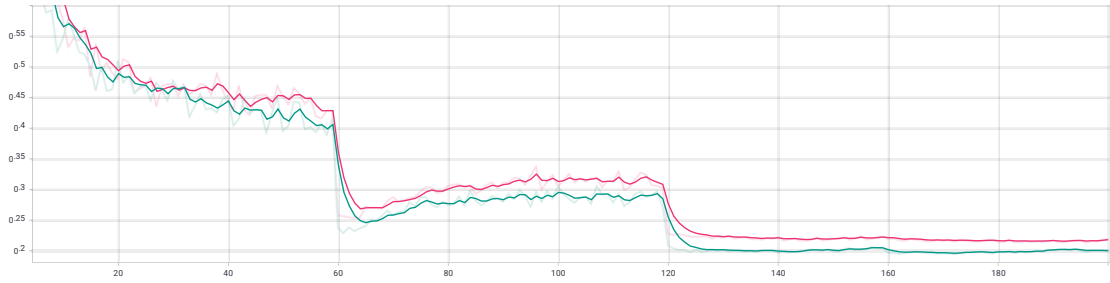
Fig.5. SE-ResNet101(red) SE-R-ResNet101(green)

# 5 Ablation Study

## 5.1 How much does auxiliary classifiers contribute to the final accuracy

As we know, auxiliary classifier itself is a way to improve accuracy, so how much of the improvement shown in Table 3 comes from auxiliary classifier? We added the same auxiliary classifier used in the corresponding SE-R-ResNet to SE-ResNet, which we will refer to as SE-A-ResNet. The other conditions of the experiment, such as training strategy, lossW, etc., are kept as the original settings. The comparison results are shown in Table 4. It can be seen that using auxiliary classifiers on SE-ResNet still has a good effect of improving accuracy, but the results are still lower from SE-R-ResNet, which proves that the modifiction is effective.

| Net | Accuracy | Net | Accuracy | Net | Accuracy |
| --- | --- | --- | --- | --- | --- |
| SE-ResNet34 | 78.12 | SE-A-ResNet34 | 78.78 | SE-R-ResNet34 | 78.98 |
| SE-ResNet50 | 78.22 | SE-A-ResNet50 | 79.25 | SE-R-ResNet50 | 79.96 |
| SE-ResNet101 | 78.54 | SE-A-ResNet101 | 79.86 | SE-R-ResNet101 | 80.48 |

Table4

## 5.2 Effect of dropout ratio in auxiliary classifier

As described in section4, the experimental default auxiliary classifier uses a dropout rate of 0.7. Considering that SE-ResNet itself has a different network structure than GoogLeNet and is used for a different purpose, it is necessary to experimentally test other cases to see if there is a better choice. We still tested on cifar100, using SE-R-ResNet50 as an example, and the results are shown in Table 5. It can be seen that the highest accuracy is achieved when the dropout rate is taken as 0.7, but in general, the impact of this parameter is not very significant.

| Dropout ratio | Accuracy |
| --- | --- |
| 0.3 | 79.78 |
| 0.5 | 79.93 |
| 0.7 | 79.96 |
| 0.9 | 79.77 |

Table5

## 5.3 Effect of auxiliary classifier position

SE-ResNet50 has a total of four stages, to explore the effect of adding auxiliary classifier at different locations, we tested the attention network by connecting an auxiliary classifier after the first three stages respectively, and then introducing the results into the later stage, and the results are shown in Table 6.

It can be found that for SE-Net, using auxiliary classifier near the input layer works better compared to adding it near the output layer. Notice that the best result of adding in middle exceeds the best result of adding two auxiliary classifiers before. This may be because the adjustment of the lossW of the two aux is not fine enough.

According to the results here, for SE-ResNet, the theoretical best way to add two auxiliary classifiers is to add them after the first and second stages.

| Position | Accuracy |
|---|---|
| End | 78.42 |
| Middle | 80.12 |
| Front | 78.96 |

Table6 End means add aux after stage3, Middle means add aux after stage2, Front means add aux after stage1

## 5.4 Effect of lossW

The accuracy rates corresponding to different values of lossW taken for the three networks SE-R-ResNet 34, 50, and 101 are shown in Tables 7, 8, and 9. It can be found that among the lossW of all the tests of the three networks of SE-R-ResNet, the values of lossW that obtain the best results are not the same for each network(0.4, 0.4 in SE-R-ResNet34, compared to 0.3, 0.3 in SE-R-ResNet50 and 0.3, 0.2 in SE-R-ResNet101). By observing and analyzing this result, roughly speaking, it would be more appropriate to add auxiliary classifier in the layer farther away from the input layer with relatively smaller lossW. The difference between SE-R-ResNet34 and 50 is that the two layers of convolution in each block are changed to three layers, and the overall relative structure does not change, so the lossW is uniformly reduced by 0.1. However, SE-R-ResNet101 has a significant increase in the number of layers in the third stage compared to SE-R-ResNet50, and the number of layers between the two auxiliary classifiers is also very large, and it is not appropriate to use the same lossW1 and lossW2 again at this time, Therefore, we tested the combinations that are more likely to be the best choice with reference to the results in Tables 7 and 8. The basic principle is to first fix lossW1 as 0.3, change lossW2 to arrive at the best choice of lossW2, and then vice versa to arrive at the best choice of lossW1.

| lossW1,W2 | 0, 0(SE-ResNet) | 0.3, 0.3 | 0.4, 0.4 | 0.5, 0.5 | 0.6,0.6 |
|---|---|---|---|---|---|
| Accuracy | 78.12 | 78.46 | 78.98 | 78.68 | 78.59 |

Table7 SE-R-ResNet34

| lossW1,W2 | 0, 0(SE-ResNet) | 0.2, 0.2 | 0.3, 0.3 | 0.4, 0.4 | 0.5, 0.5 | 0.6,0.6 | 0.7, 0.7 |
|---|---|---|---|---|---|---|---|
| Accuracy | 78.22 | 79.26 | 79.96 | 79.69 | 79.16 | 79.58 | 79.31 |

| lossW1,W2 | 0, 0(SE-ResNet) | 0.3, 0.1 | 0.3, 0.2 | 0.3, 0.3 | 0.4, 0.2 | 0.2, 0.2 |
|---|---|---|---|---|---|---|
| Accuracy | 78.54 | 80.09 | 80.48 | 80.04 | 79.66 | 78.89 |

Table9 SE-R-ResNet101

## 5.5 Effect of reduction ratio

The role of r in SE-ResNet is to control the number of parameters in the SE Block, the smaller the r the larger the number of parameters, generally speaking, will achieve better results, here we also take SE-R-ResNet50 as an example to see the impact of different values of r on the accuracy rate, the results are shown in Table 10. We take r=8 by default.

| r | 8 | 12 | 16 |
|---|---|---|---|
| Accuracy | 79.96 | 79.69 | 79.53 |

Table10 SE-R-ResNet50

## 5.6 Replace auxiliary classifier with auxiliary model

As mentioned before, the main idea of the improvement of this paper is to get the output result in advance to introduce it into the attention block as a part of the input, then the method of introducing the results in addition to auxiliary classifier can also consider using auxiliary model. Specifically: the image data are simultaneously fed into the auxiliary model, and the output is passed into SENet's attention network, combined in a similar way to auxiliary classifier. We have also done related experiments to explore this possibility. The specific setup of the experiment is as follows: the auxiliary model is chosen from GoogLeNet and its loss is multiplied by a weight lossW added to the loss of the main network, which is modified from SE-ResNet50. Table 11 shows the results obtained by taking different lossW.

| lossW | 0(SE-ResNet) | 0.1 | 0.2 | 0.3 | 0.4 | 0.5 |
|---|---|---|---|---|---|---|
| Accuracy | 78.22 | 77.27 | 77.05 | 78.05 | 77.32 | 77.61 |

Table11 SE-R-ResNet50

According to the results in the table, the method has not improved. However, we can see that 78.05 is close to the original result, and if we use a finer lossW, the method may still make some progress. This part is left for future study.

# 6 Effect Of Introducing The Result Into The Attention

According to Figure 5, we can know that introducing the prediction results into the attention network is indeed effective, so why is it effective to do so? We believe that feeding the prediction results into the attention network in advance helps the attention network to give a better distribution of channel attention weights based on the prediction results. For example, to identify a 'car', if the 'car' has a high probability in the prediction results obtained in advance, then the attention network naturally tends to give more weight to the channel where the features with high 'car' relevance are located, while reducing the weight for other categories with small probability. This reduces the interference from other categories and reduces the difficulty of classification.

We did an experiment to prove this conjecture, and we used the middle network in

section5.3 as an example to compare with the original SE-ResNet, with 128 randomly selected inputs from cifar100. The y-axis is the standard deviation of s given by all SE modules in stage3. The results are shown in Figure 6. It can be found that the improved network greatly improves the discrimination of attention between channels, and this confirms our conjecture.

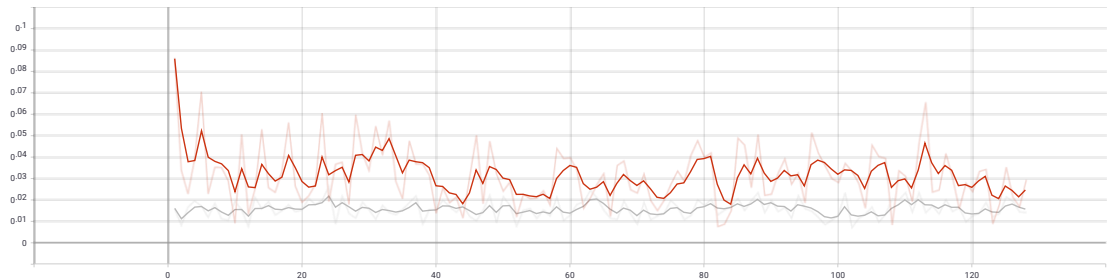

Fig.6. The red line is the layer with the lowest mean standard deviation of the modified network, and the gray line is the layer with the highest mean standard deviation of the original network. For brevity, the standard deviations of other layers are not presented here

## 7 Conclusion

In this paper, we propose a new attention modification method that allows some existing attention networks to be improved even further by introducing prediction results. Although this paper only does some experiments on classification networks and only on SE-Net, the method can also be applied in areas such as target detection and on other networks, and we believe that some progress could be made as well.